\documentclass[journal]{IEEEtran}
\IEEEoverridecommandlockouts

\usepackage{cite}
\usepackage{amsmath,amssymb,amsfonts}
\usepackage{algorithmic}
\usepackage{float}
\usepackage{booktabs}
\usepackage{enumitem}
\usepackage{tabularx}
\usepackage{makecell}
\usepackage{array}
\usepackage{graphicx}
\usepackage{textcomp}
\usepackage{xcolor}
\usepackage[letterpaper]{geometry}
\usepackage{comment}

\begin{document}

\title{Self Capacitive Tactile Sensor System designed for Companion Robots}

\author{
    Mohsin Ali$^{1}$,
    Hidenobu Sumioka$^{2}$,
    Shuhei Ikemoto$^{1}$
\thanks{This work was supported by JSPS KAKENHI Grant Number 25H02619 and JP25H01161.}
\thanks{$^{1}$ M. Ali and S. Ikemoto are with the Graduate School of Life Science and Systems Engineering, Kyushu Institute of Technology, 2-4 Hibikino, Wakamatsu, Kitakyushu, Fukuoka 808-0196 Japan. (e-mail:  {\tt\small ikemoto@brain.kyutech.ac.jp})  }
\thanks{$^{2}$ S. Sumioka is with Advanced Telecommunications Research Institute International, 2-2-2 Hikaridai Seika-cho, Sorakugun, Kyoto
619-0288 Japan.}
}

\maketitle
\begin{abstract}
Tactile sensing is essential for humanoid robots to achieve safe physical interaction, dexterous manipulation, and truly human-like responsiveness. However, the design of such systems remains challenging. Conventional approaches often suffer from complex multilayer structures, intricate wiring, high cost, and poor scalability, making it difficult to realize full-body tactile sensing with real-time, low-latency detection while maintaining minimal computational load on the robot’s main processor. In this work, we present a simple, scalable and hardware friendly tactile sensing system for a companion humanoid robot based on the self-capacitance principle. The proposed sensor system employs a single conductive fabric layer with a conductive fabric wire architecture and does not require intricate electrode patterning. Scalability was demonstrated by fabricating a 100-point sensor array on a flexible printed circuit (FPC). Evaluation across sampling frequencies showed that 10 Hz is insufficient and misses transient events, whereas 100 Hz and 1000 Hz reliably capture and clearly distinguish all interaction types: gentle touch, slow tapping, fast tapping, and hitting. A decision-tree classifier was implemented directly on the FPGA, offloading real-time inference from the Raspberry Pi 4 with minimal latency and negligible power overhead. This design fully meets the tactile sensing requirements of the HIRO-chan robot and is well-suited for full-body tactile sensing in HIRO-chan and other companion robots.
\end{abstract}


\section{Introduction}
Engaging in close interactions with humans and environments in daily life is one of the most challenging goals in robotics. Tactile sensing is the most direct approach to capture the physical interactions that occur there \cite{art2,art3}. Various methods have been proposed including capacitive \cite{art6,art6.1}, resistive\cite{art7,art7.1}, piezoresistive\cite{art8,art8.1}, and optical sensing\cite{art9,art9.1} to give robots tactile perception each offering trade-offs in sensitivity, response time, power consumption, and fabrication complexity.

Among these approaches, capacitive sensors have attracted considerable attention due to their relatively simple structures, excellent linearity, high sensitivity, low power consumption, fast dynamic response\cite{art10,art11,art12.1}, and unique capability to detect both direct contact and proximity\cite{art13,art14}. These characteristics make it a promising solution to achieve practical whole-body tactile perception in soft robots.
\begin{figure}[t]
\centerline{\includegraphics[width=0.5\textwidth]{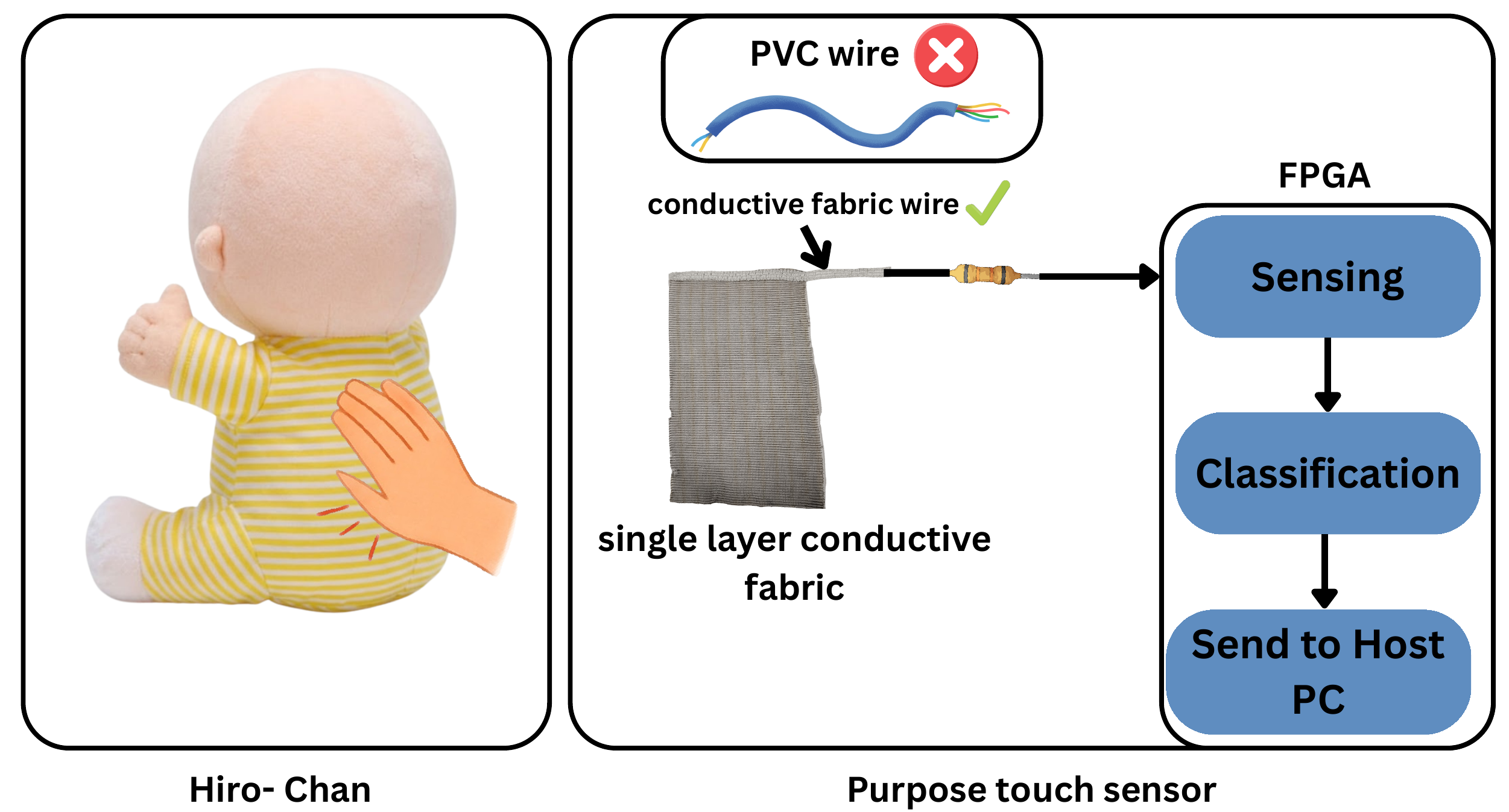}}
\caption{System overview of the proposed tactile sensing architecture for the Hiro-chan robot, illustrating a soft, fabric-based capacitive sensor using conductive textile wiring and FPGA-based signal acquisition.}
\label{fig:Figure 4scheeet}
\end{figure}
A considerable amount of work has been done on developing full-body tactile sensor systems for robots that address conformability, scalability, and hardware simplicity. Notable examples include piezoresistive approaches, such as layered textile-based structures using a single piezoresistive textile layer with distributed electrodes for soft whole-body tactile skin via electrical resistance tomography (ERT)\cite{art15}, and scalable fabric tactile arrays in row-column matrix configurations for affordable high-resolution sensing\cite{art16}.
Other designs include capacitive or matrix-based flexible arrays, as seen in large-area high-resolution skin-inspired sensors that achieve detailed tactile perception, often with multi-layer constructions for enhanced resolution and durability\cite{art17}. Stretchable textile-based electronic skins with reduced wiring complexity have also been developed for large-area coverage, detecting both contact position and force in real time while accommodating complex robot surfaces\cite{art18}. Modular approaches feature full-body e-skin patches for tactile gesture recognition, typically involving modular assemblies that allow customizable coverage without excessive wiring\cite{art19}.
The above approaches rely on multiple layers and involve complex fabrication processes, which may not be suitable for companion robots such as HIRO-chan a child robot designed for easy adaptability, safe interactions, and therapeutic engagement with dementia patients\cite{art24,art25,art26}.
One suitable approach for the tactile sensor system is a flexible fabric-based tactile sensor system \cite{art27}. However, its major drawback is that it requires a microcontroller for each sensor and complex wiring four wire per sensor(power,GND,SDA SCL), which limits scalability and introduces rigid components that may make the system uncomfortable for users of companion robots such as HIRO-chan.

In this work, we propose a tactile sensor system based on the self-capacitance principle. It requires only a single conductive layer such as conductive fabric along with flexible conductive fabric wire to connect the sensors to the sensing unit and does not require a microcontroller with each sensor and simple wiring as shown in Figure \ref{fig:Figure 4scheeet}. This makes it a simple and scalable approach and makes it highly suitable for soft child companion robots like HIRO-chan. Furthermore, the system is computationally efficient, as the classification of different interaction types is performed directly on an FPGA. This offloads processing from the main microcontroller and makes the solution well-suited for resource-constrained edge devices.

\begin{figure}[t]
\centerline{\includegraphics[width=0.5\textwidth]{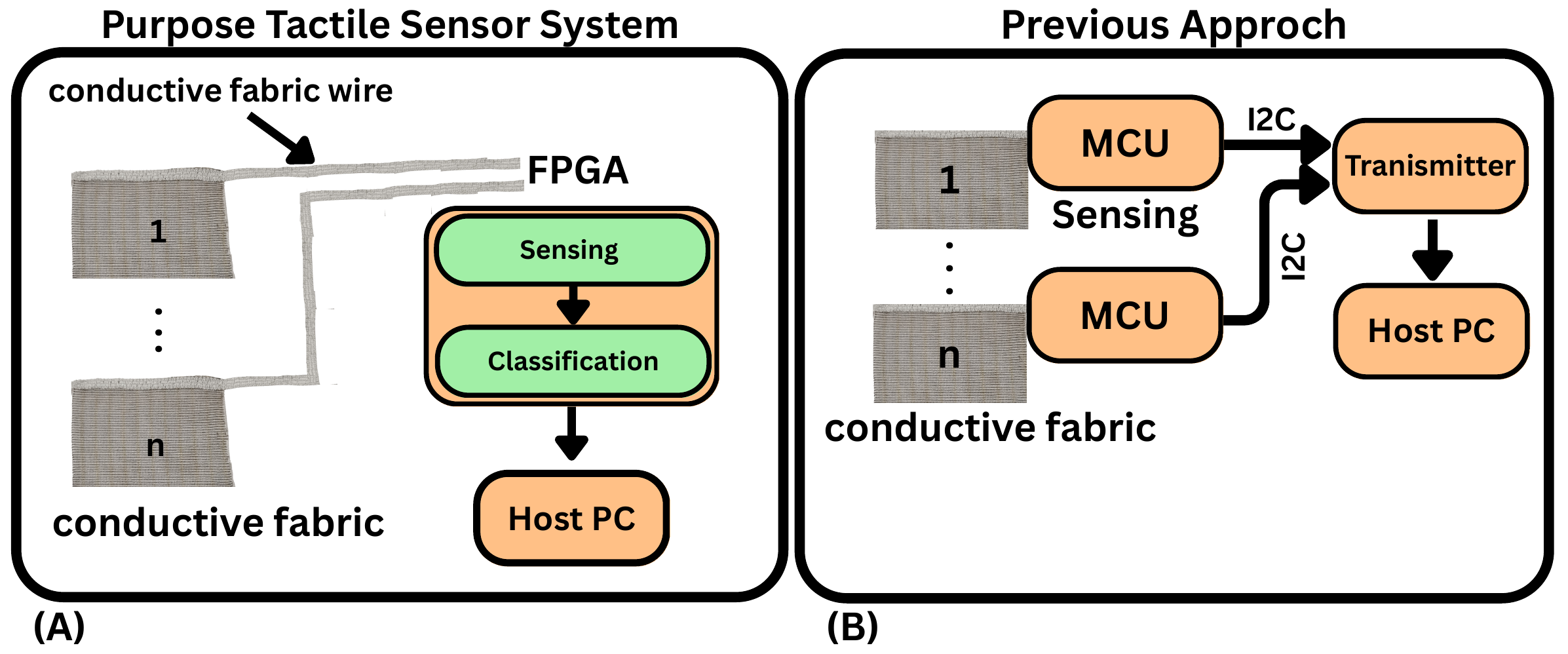}}
\caption{Comparison between the previous approach \cite{art27} and the proposed tactile sensor system.(A) our purpose tactile sensor system (B) Previous Approach}
\label{fig:Figure 4refff}
\end{figure}

\section{Design Challenges}
The Hiro-chan robot is a companion humanoid robot platform developed for therapeutic interactions with dementia patients. Hiro-chan is designed to mimic a young child in appearance and behavior, featuring soft, approachable materials and simple expressive capabilities to evoke empathy and reduce anxiety in elderly users. Its primary functions include basic conversational responses, gesture recognition, and companionship activities, which have shown promise in clinical studies to improve cognitive engagement and emotional well-being in dementia care.
\begin{figure}[t]
\centerline{\includegraphics[width=0.5\textwidth]{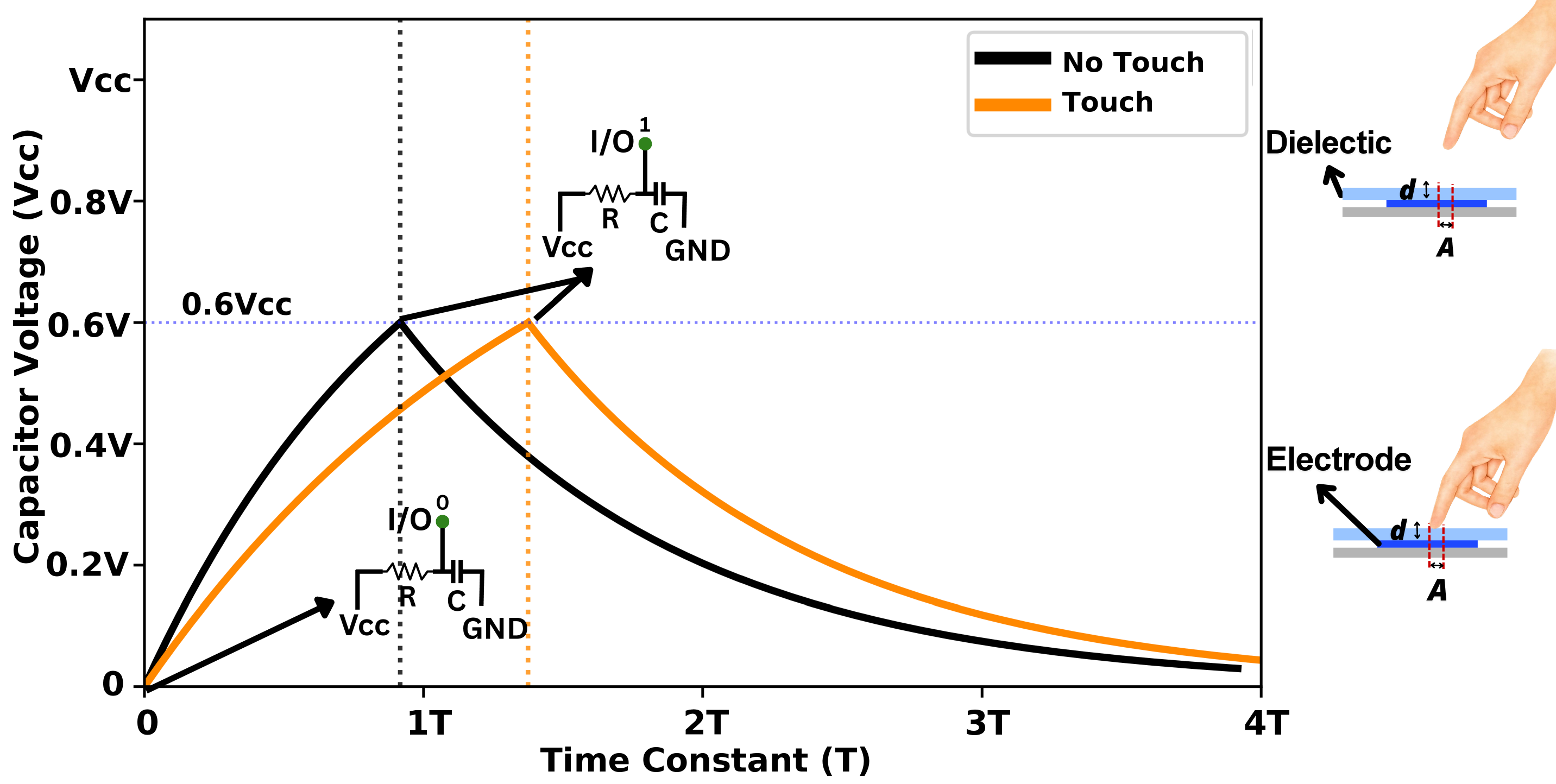}}
\caption{Capacitor voltage as a function of time constant for touch and no-touch states. In the no-touch condition, the RC time constant required to reach the 0.6Vcc where the digital input transitions from LOW to HIGH is smaller than that of the touch state, indicating increased capacitance during touch events}
\label{fig:Figure 4sense}
\end{figure}
In this work, we expand the Hiro-chan capabilities by incorporating advanced touch sensing, enabling more intuitive physical interactions. There are various design criteria for touch sensors system in the context of human–robot touch interactions \cite{art28}.
\begin{enumerate}
    \item  The appearance and feel of the skin should be appropriate to the robot’s appearance and behavior, given that an unexpected sensation is likely to disrupt any social connection that has been developed
    \item  Electrical wiring and terminations should be compatible with skin stretch and be kept to a minimum.
    \item Electronics should be easily adjustable to different hardware platforms, designed for minimum power consumption and be suitable for battery-powered operation.
    \item All data processing, touch interpretation and decision-making should be done in real time with minimum latency.
\end{enumerate}
Considering the design constraints outlined above that are not satisfied by existing approaches \cite{art27}, we propose a self-capacitance-based tactile sensor system for full-body sensing in the Hiro-chan robot. The design uses a single conductive-fabric electrode and conductive-fabric wire, preserving the robot's soft feel and appearance while eliminating rigid surface components such as microcontrollers and stiff wiring. As shown in Figure \ref{fig:Figure 4refff}, the proposed approach is simpler and less bulky than the previous design, requiring only a single conductive-fabric wire instead of four wires per sensor (power, ground, SDA, and SCL) for operation and communication.
For sensing and processing, the Lattice ICE40HX8K FPGA breakout board is selected due to its excellent power efficiency and fast sensing capability. To improve computational efficiency for edge devices, a machine learning model is deployed directly on the FPGA. This approach offloads processing from the main microcontroller, making the system hardware friendly, well-suited for resource-constrained environments and enabling real-time sensing and decision-making.

\section{Sensing Principle}

To enable responsive touch interaction in Hiro-chan, a capacitive tactile sensor was employed due to its fast response and relatively simple design. In particular, the system uses self-capacitance sensing because it offers a fast response rate, a simple circuit structure, and a single-electrode configuration, making it scalable and easy to implement. The sensing principle in this system is based on a charge-and-measure RC relaxation method for self-capacitance detection. A fixed resistor in series with the variable electrode capacitance, where the total capacitance consists of the inherent (parasitic) capacitance of the electrode plus any additional capacitance of a nearby finger or conductive object. When a finger approaches, it forms a parallel capacitive path to ground through the body, increasing the overall capacitance at that electrode.This change can be modeled using the parallel plate capacitance equation
\begin{equation}
C = \varepsilon_r\varepsilon_0 \frac{A}{d}
\end{equation}
where $  C  $ is the capacitance in farads, $  \epsilon_0  $ is the permittivity of free space (a constant approximately equal to $  8.85 \times 10^{-12}  $ F/m. $  A  $ is the effective overlapping area between the electrode and the finger and $  d  $ is the distance separating them as the finger approach the d decrease and as a results increase in the capacitance.

\begin{figure}[t]
\centerline{\includegraphics[width=0.5\textwidth]{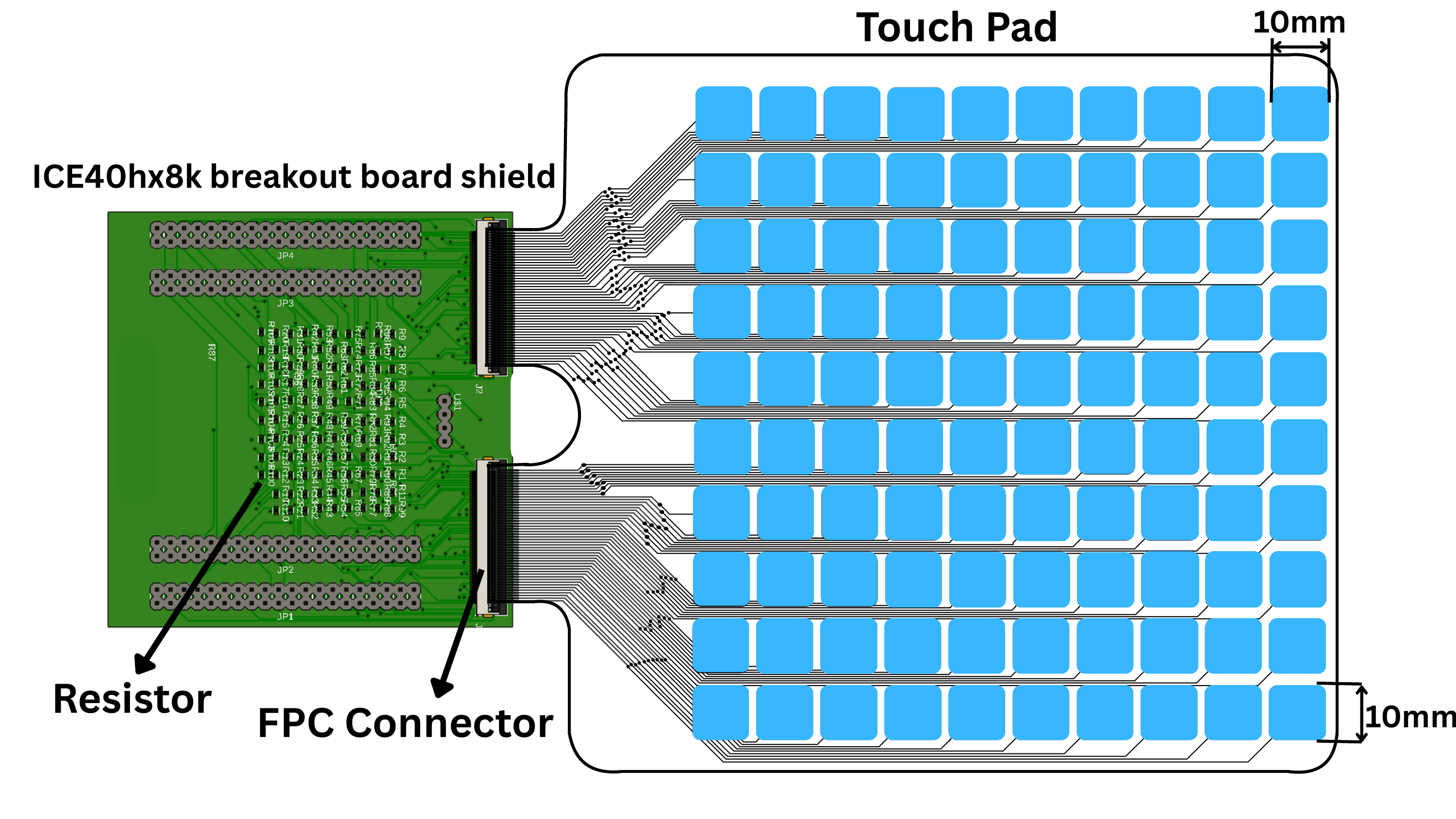}}
\caption{Overall system architecture comprising a custom PCB shield interfaced with an FPGA and a flexible printed circuit incorporating 100 capacitive touch points.}
\label{fig:Figure 4sys}
\end{figure}

The electrode is charged to Vcc through the fixed resistor, with the voltage across the electrode increasing exponentially during charging according to the equation 
\begin{equation}
     V_c(t) = V_{cc} \left(1 - e^{-t / (R C)}\right)  
\end{equation}
as shown in Figure \ref{fig:Figure 4sense}. Where $  V_c(t)  $ is the voltage at time $t$, $R$ is the resistance, and $  C  $ is the capacitance. The FPGA then measures the time until the digital pin toggles from logic 0 to logic 1 using a fast polling loop. As the voltage on the electrode increases during the charging phase and reaches approximately 0.6 × VCC (the typical logic high threshold for CMOS input), the digital pin switches from logic 0 to logic 1. This variation in the charging time is detected by precisely counting the number of FPGA clock cycles from the start of the charging process until the 0-to-1 toggle occurs. For discharging, the pin is set to high impedance mode to discharge the electrode, where the voltage decays exponentially as
\begin{equation}
   V_c(t) = V_{cc} \, e^{-t / (R C)}.
\end{equation}
 A higher capacitance due to finger approach or touch causes a slower charging rate, resulting in more clock cycles compared to the no-touch baseline state. This difference allows for reliable detection of self-capacitance changes across all 100 sensing points in real time.
\begin{figure}[b]
\centerline{\includegraphics[width=0.5\textwidth]{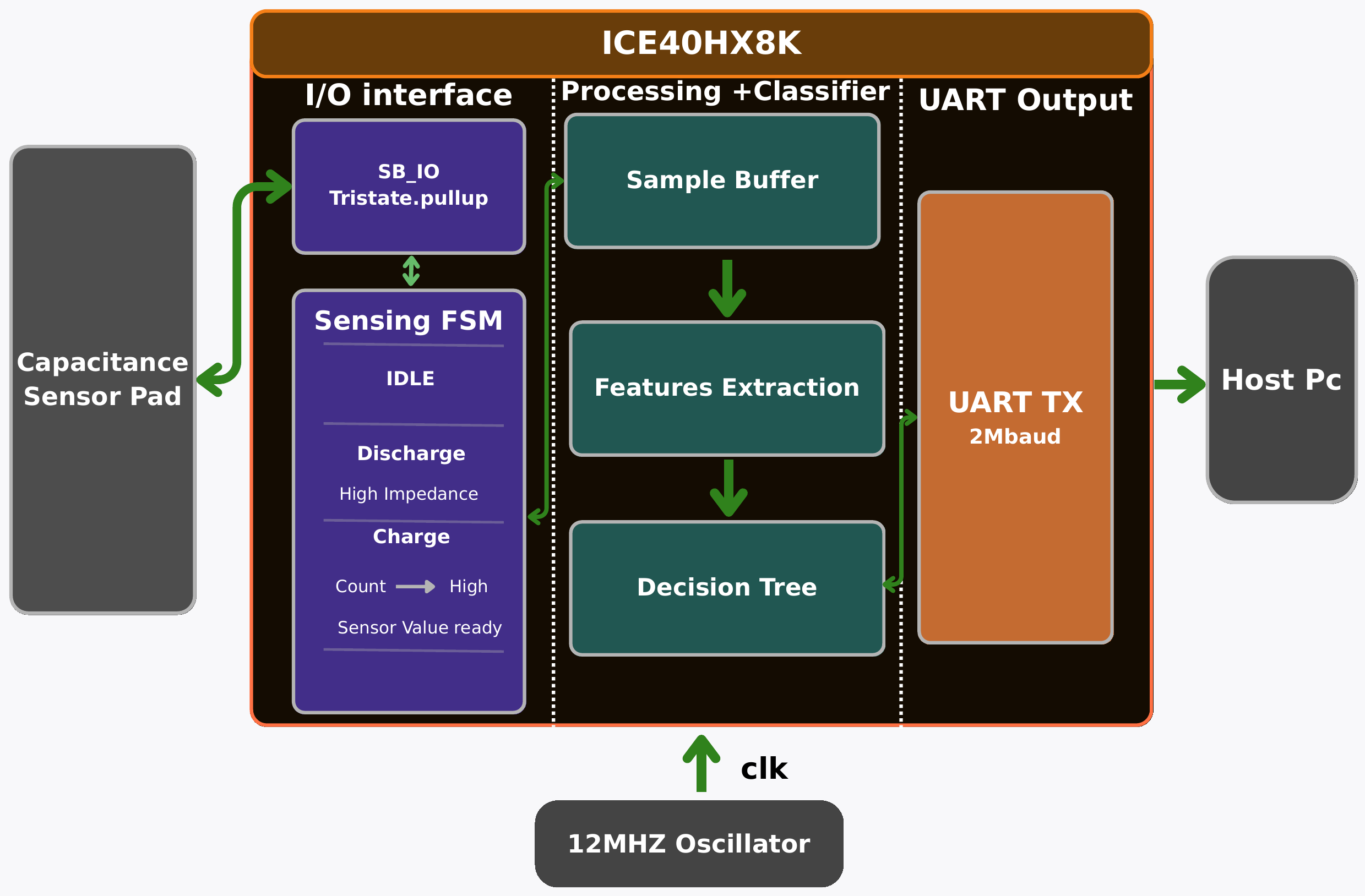}}
\caption{Hardware architecture of the real-time capacitance sensing and tapping classification system implemented on ICE40HX8K FPGA, detailing the sensor interface, finite state machine, processing pipeline, and 2 Mbaud UART output.}
\label{fig:Figure arc}
\end{figure}

\section{System Design}
To demonstrate the scalability of the proposed tactile sensing system, a 100-sensor flexible PCB was design and used, as shown in Figure \ref{fig:Figure 4sys}. The proposed tactile sensing system consists of three key components.
\begin{enumerate}[label=\Alph*.]
    \item FPGA ICE40hx8k
    \item Custom PCB board for ice40hx8k sheild
    \item sensing sheet 
\end{enumerate}

\subsection{FPGA ICE40hx8k}
\begin{figure*}[t]
\centerline{\includegraphics[width=\textwidth]{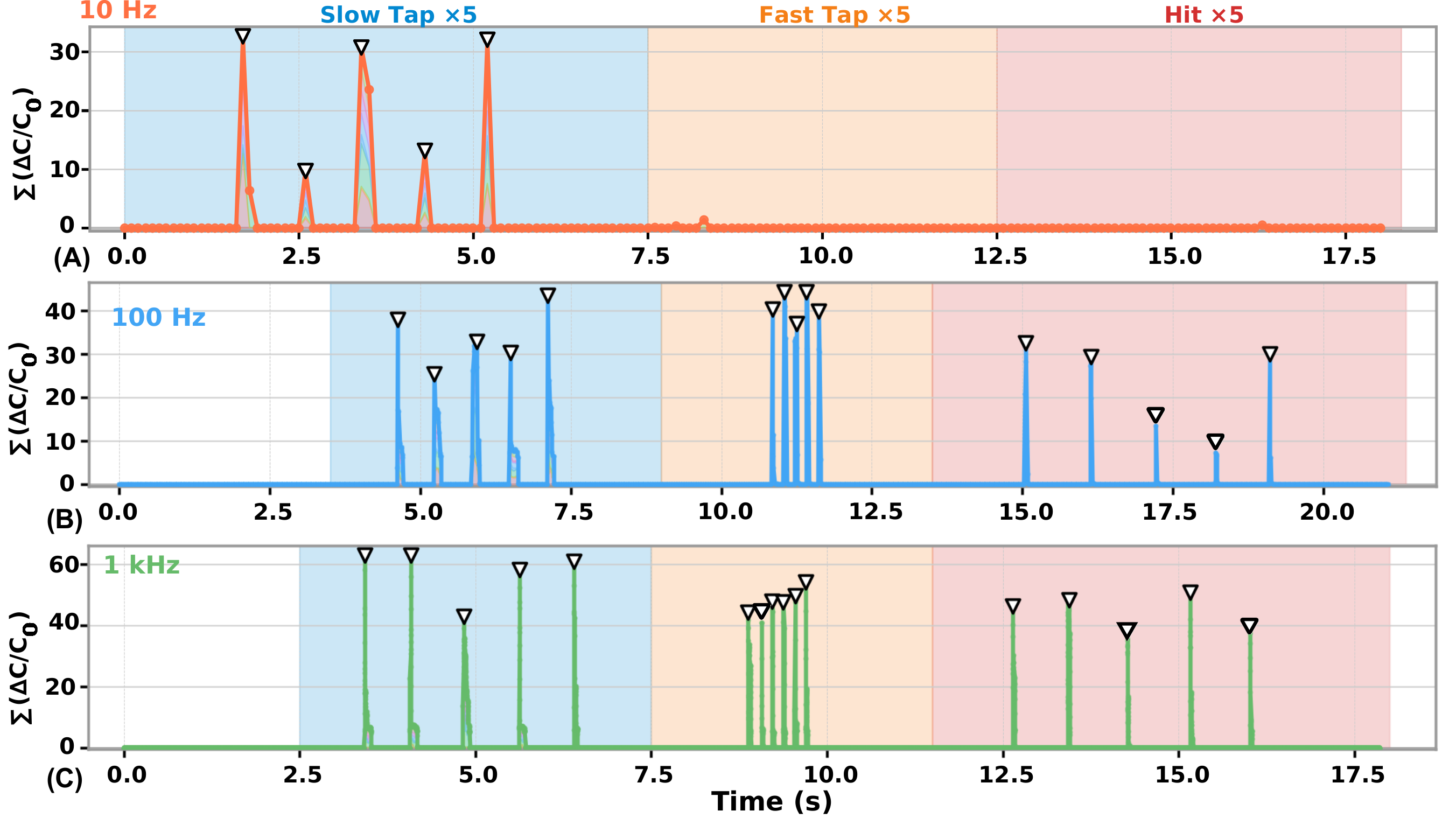}}
\caption{Comparison of the summed relative capacitance change $\sum \left( \frac{\Delta C}{C_0} \right)$ captured at different sampling frequencies in response to slow tapping (×5), fast tapping (×5), and hitting (A) at 10hz,(B) at 100hz and (C) 1khz }
\label{fig:Figure cla}
\end{figure*}

One of the main units in the tactile sensing hardware is the Lattice iCE40 HX8K FPGA mounted on the CT256 breakout board, which serves as the core processor responsible for high-speed self-capacitance sensing and real-time classification of interaction types (no touch,touch,slow tapping, fast tapping, and hitting). The complete system architecture implemented on this FPGA, including the I/O interface, Sensing FSM, processing-classifier pipeline, and UART output, is shown in Figure~\ref{fig:Figure arc}. This compact FPGA device integrates 7,680 logic cells (each built around a 4-input look-up table and a flip-flop), 128 kbits of embedded block RAM, and dual phase-locked loops (PLLs) that provide flexible clock multiplication and synthesis from the board’s stable 12 MHz onboard crystal oscillator. Using the parallel architecture of FPGA and the low resource overhead, the implemented capacitance measurement logic, charge-transfer circuitry, and decision-tree classifier achieve deterministic response times, allowing reliable real-time detection of subtle capacitance variations and accurate classification of different tapping interactions. This capability makes the system a complete, hardware-friendly, low-power FPGA-based capacitive tactile sensor system with gesture recognition, particularly suitable for battery-powered companion robots that require responsive touch sensing combined with gesture discrimination.

\subsection{Custom PCB board}
One of the main units in the experimental setup is the custom 4-layer PCB shield designed to mount directly on the Lattice iCE40 HX8K CT256 breakout board. This shield includes 100 SMD(surface mounted resistor) resistors of 10\,M$\Omega$ forming 100 separate RC circuits for self-capacitance sensing at 100 points, enabling the connection of up to 100 sensors. For this setup, It also features two 50 pins FPC connectors to easily attach flexible PCBs carrying the sensing electrodes. The 4-layer design helps maintain good signal quality and low crosstalk between channels.

\subsection{Sensing Sheet}
To evaluate the scalability of the proposed sensing method, the flexible printed circuit (FPC) board was designed and fabricated using a two-layer design that connects to the custom PCB shield via FPC connectors. The top layer (facing the user or overlay) contains the 100 touch sensor electrodes for direct detection of self-capacitance changes, while the sensing lines and routing traces are placed on the back (bottom) layer. This configuration prevents false touches by shielding the sensitive traces from unintended contact or proximity effects, similar to routing traces on the underside of a board to avoid interference. Overall, this layered approach improves reliability in multi-point capacitive sensing applications.

\section{Experimental Results and Discussion}
To evaluate the performance of the capacitive touch sensing system at different temporal resolutions, experiments were conducted at three sampling frequencies: 10 Hz, 100 Hz, and 1 kHz. Three types of touch interactions were considered for classification: fast tapping, slow tapping, and hitting.

At each sampling frequency, a total of 15 touch events were performed. Figure \ref{fig:Figure cla} shows the corresponding capacitance time-series signals acquired at 10 Hz, 100 Hz, and 1 kHz. The purpose of this experiment was to evaluate whether the selected sampling frequencies provide sufficient temporal resolution to capture the characteristic features of each interaction type. The results demonstrate that sampling frequency has a significant impact on the detection and representation of touch events. Due to their longer duration, slow tapping events are reliably detected at all three sampling frequencies. In contrast, fast tapping and hitting events are adequately resolved only at 100 Hz and 1 kHz. At 10 Hz, the limited temporal resolution fails to capture short-duration events such as hitting and fast tapping, resulting in the loss of critical interaction data in human-robot interaction.

Although 1 kHz sampling yields the highest temporal resolution and the most accurate representation of signal waveforms, it generates a significantly higher data rate. For a robot system relying on an edge computing device, processing and handling the large volume of data at 1 kHz in real time poses considerable challenges in terms of computational load, memory usage, and power consumption. Consequently, 100 Hz represents the most practical sampling frequency: it successfully captures both fast tapping and hitting events with no noticeable degradation of essential information while remaining fully feasible for edge-device deployment.

Classification of the three interaction types (slow tapping, fast tapping, and hitting) was performed on the basis of three signal features: event duration, peak amplitude, and inter-event interval. Hitting events are characterized by isolated high-amplitude impulses with short duration.

\begin{figure}[t]
\centerline{\includegraphics[width=0.5\textwidth]{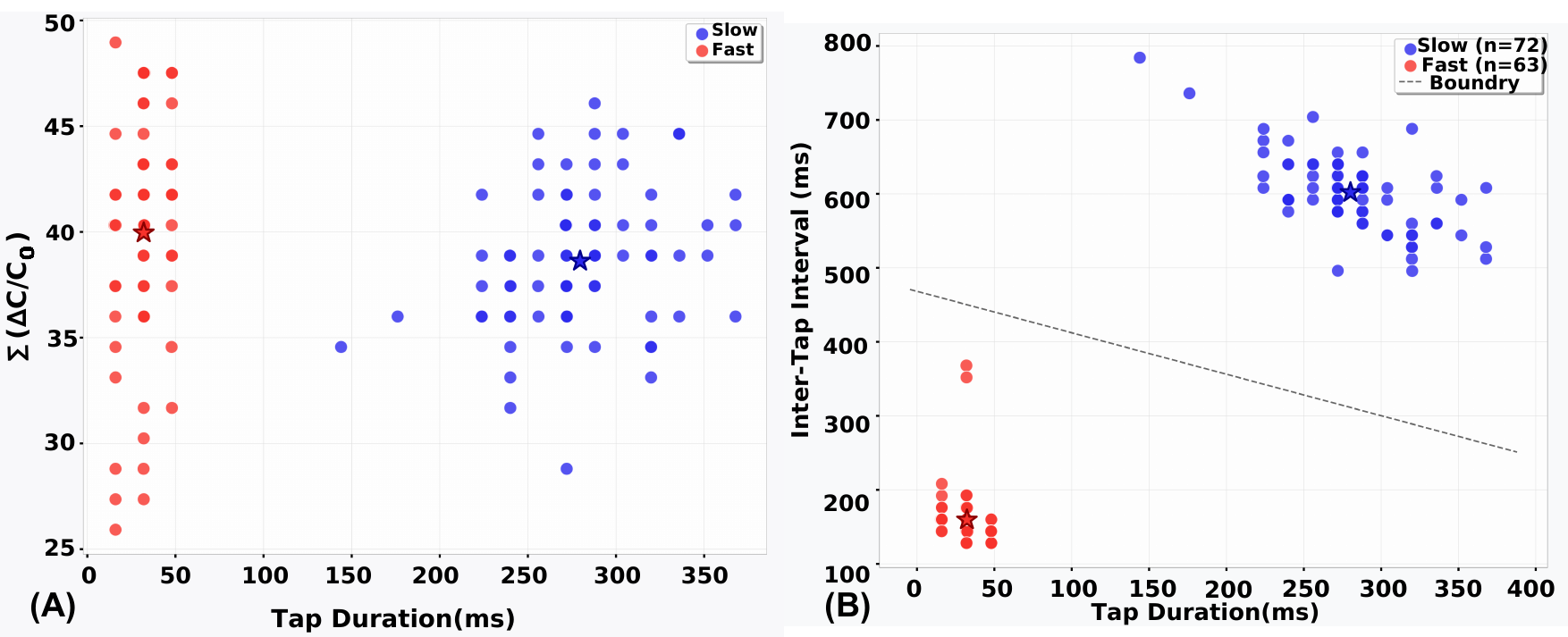}}
\caption{Classification of fast tapping and slow tapping events: (A) based on amplitude versus time duration and (B) classification based on inter-tap interval versus tap duration.}
\label{fig:Figure inte}
\end{figure}

The most challenging cases for classification are the slow tapping and fast tapping events. As shown in Figure~\ref{fig:Figure inte}(A), slow and fast tapping events are clearly separated in the scatter plot of event duration versus inter-event interval. Figure~\ref{fig:Figure inte}(B) offers further discrimination through the relationship between the event duration and the peak amplitude. The statistical analysis presented in Table \ref{tab:statistical_analysis}, based on the Mann-Whitney U test, further confirms this separation, revealing highly significant differences in the two signal temporal features, tap duration and inter-tap interval, between the two classes(both $p < 0.001$), with  large effect sizes. In contrast, the relative capacitance change ($\Delta C/C_0$) exhibits only a small difference between the two classes. These results indicate that temporal characteristics are the primary discriminative features and strongly support the accurate and reliable classification of slow and fast tapping events.

The main processing unit of the Hirochan robot is a Raspberry Pi 4. With 100 sensors each generating 100 readings per second, the system imposes a heavy computational load on the Raspberry Pi. Executing the classification algorithm directly on the Raspberry Pi noticeably degrades its overall performance. To overcome this limitation and make this system hardware friendly, touch-event classification is performed entirely within the FPGA, which offers both low power consumption and high processing efficiency. This edge-computing approach is considered best practice for resource-constrained robotic platforms such as the Hirochan.
\begin{table}[t]
\centering
\caption{Statistical comparison between slow and fast taps}
\label{tab:statistical_analysis}
\setlength{\tabcolsep}{2pt}
\renewcommand{\arraystretch}{1.0}
\begin{tabular}{lcccc}
\hline
\textbf{Feature} & \textbf{Slow} & \textbf{Fast} & \textbf{p} & \textbf{Cohen’s d} \\
\hline
Tap Duration (ms) &
279.6 $\pm$ 41.5 &
32.0 $\pm$ 10.6 &
$<0.001$ &
8.17 \\
Inter-Tap Interval (ms) &
602.0 $\pm$ 52.1 &
159.5 $\pm$ 40.3 &
$<0.001$ &
9.50 \\
$\Delta C/C_0$ &
38.6 $\pm$ 3.3 &
40.0 $\pm$ 5.7 &
0.021 &
0.29 \\
\hline
\end{tabular}
\end{table}
The FPGA employed is the compact Lattice ICE40 HX8K, which provides only limited resources (approximately 7,640 LUTs). Deploying a classification model on this device is challenging due to its constrained logic capacity. To overcome this limitation, three candidate machine learning algorithms were evaluated for real-time deployment on the resource-constrained iCE40 HX8K FPGA based on two key performance metrics: classification accuracy and hardware resource utilization (measured in LUTs). The comparison results are summarized in Table~\ref{tab:algorithm_comparison}. The entire classification task is offloaded to the FPGA, significantly reducing the processing burden on the Raspberry Pi 4 while maintaining real-time performance.
Currently, the FPGA classifies touch events into five distinct categories: no touch, touch, slow tapping, fast tapping, and hitting. This comprehensive on-board classification makes the system fully self-contained and ready for direct integration into any target application.
\begin{table}[htbp]
\centering
\caption{Algorithm Comparison}
\label{tab:algorithm_comparison}
\begin{tabular}{|>{\raggedright\arraybackslash}p{2.5cm}|c|c|c|c|c|}
\hline
\textbf{Algorithm} & \textbf{Accuracy} & \textbf{FPGA LUTs} & \textbf{Selected} \\
\hline
Random Forest  & 92.3\% & 15,727  & No \\
\hline
SVM (RBF Kernel) & $\sim$88\% & 12000   & No \\
\hline
\textbf{Decision Tree} & \textbf{90.4\%} & \textbf{2067}  & \textbf{YES} \\
\hline
\end{tabular}
\end{table}
As shown in Table~\ref{tab:algorithm_comparison}, the Random Forest algorithm achieves a high accuracy of 92.3$\%$ but requires approximately 15,000 LUTs. This resource demand greatly exceeds the limited capacity of the iCE40 HX8K FPGA, rendering it unsuitable for deployment. Similarly, the Support Vector Machine (SVM) only ~88$\%$ accuracy while consuming around 12,000 LUTs, which is also well beyond the practical limits of the platform.
In contrast, the Decision Tree algorithm delivers a competitive accuracy of 90$\%$ while using only 2,067 LUTs. This makes it the most suitable model for our resource-constrained FPGA, enabling efficient on-board classification. The resulting low-power self-capacitance tactile sensing system is highly suitable for integration into mobile robots such as the Hirochan.

\section{limitation and future work}
Although the proposed system demonstrates reliable real-time self-capacitance sensing and classification of five touch states (no touch, touch, slow tapping, fast tapping, and hitting) using the ICE40HX8K FPGA, several limitations remain. The current implementation supports only 100 sensors, which is insufficient for full-body tactile coverage, and recognizes a limited set of interaction gestures. To better support natural human-baby-like interaction in caregiving environments, a richer gesture vocabulary is required. Future work will focus on scaling the system to a full-body tactile skin with several hundred sensors and deploying the Hirochan robot in care facilities. Long-term studies with dementia patients will be used to analyze natural interactions and expand the gesture set, improving the robot's emotional expressiveness, responsiveness, and effectiveness as a companion for elderly care.
\section{conclusion}
In conclusion, this work successfully demonstrates a scalable, low-cost, and hardware firendly tactile sensing system for the Hiro-chan humanoid robot, based on the self-capacitance sensing principle and seamlessly integrated with a single conductive fabric layer and conductive fabric wiring architecture. To demonstrate the scalability of the proposed system, a 100-point sensor array was fabricated on a flexible printed circuit (FPC). Systematic evaluation across sampling frequencies revealed that 10 Hz proved critically insufficient, with numerous transient events entirely undetected, whereas both 100 Hz and 1000 Hz captured the full spectrum of interactions: touch, slow tapping, fast tapping, and hitting. Furthermore, a decision tree classifier was successfully deployed within the FPGA, effectively offloading real-time inference from the Raspberry Pi 4 and achieving minimum latency with negligible power overhead. Overall, we achieved a simple, fast, and scalable tactile sensor system for full-body tactile sensing.
\bibliographystyle{IEEEtran}
\bibliography{mybib}

\end{document}